\documentclass[runningheads]{llncs}

\usepackage[mobile]{eccv}

\usepackage{eccvabbrv}

\usepackage{graphicx}
\usepackage{booktabs}

\usepackage[accsupp]{axessibility}  

\usepackage[pagebackref,breaklinks,colorlinks,citecolor=eccvblue]{hyperref}

\usepackage{orcidlink}

\begin{document}

\title{Envision3D: One Image to 3D \\ with Anchor Views Interpolation} 

\titlerunning{Envision3D}

\author{Yatian Pang\inst{3} \quad
Tanghui Jia \inst{1} \quad
Yujun Shi \inst{3} \quad
Zhenyu Tang \inst{1} \quad 
Junwu Zhang \inst{1} \quad\\
Xinhua Cheng \inst{1} \quad
Xing Zhou \inst{4} \quad
Francis E.H. Tay \inst{3} \quad
Li Yuan \inst{1,2}\thanks{Corresponding author}}
\authorrunning{Y. Pang et al.}

\institute{School of Electronic and Computer Engineering, Peking University \and
PengCheng Laboratory \and
National University of Singapore \and
Rabbitpre
\\
\email{yatian\_pang@u.nus.edu; yuanli-ece@pku.edu.cn}}

\maketitle

\vspace{-1cm}
\begin{figure}[ht]
\centering
\includegraphics[width=\linewidth]{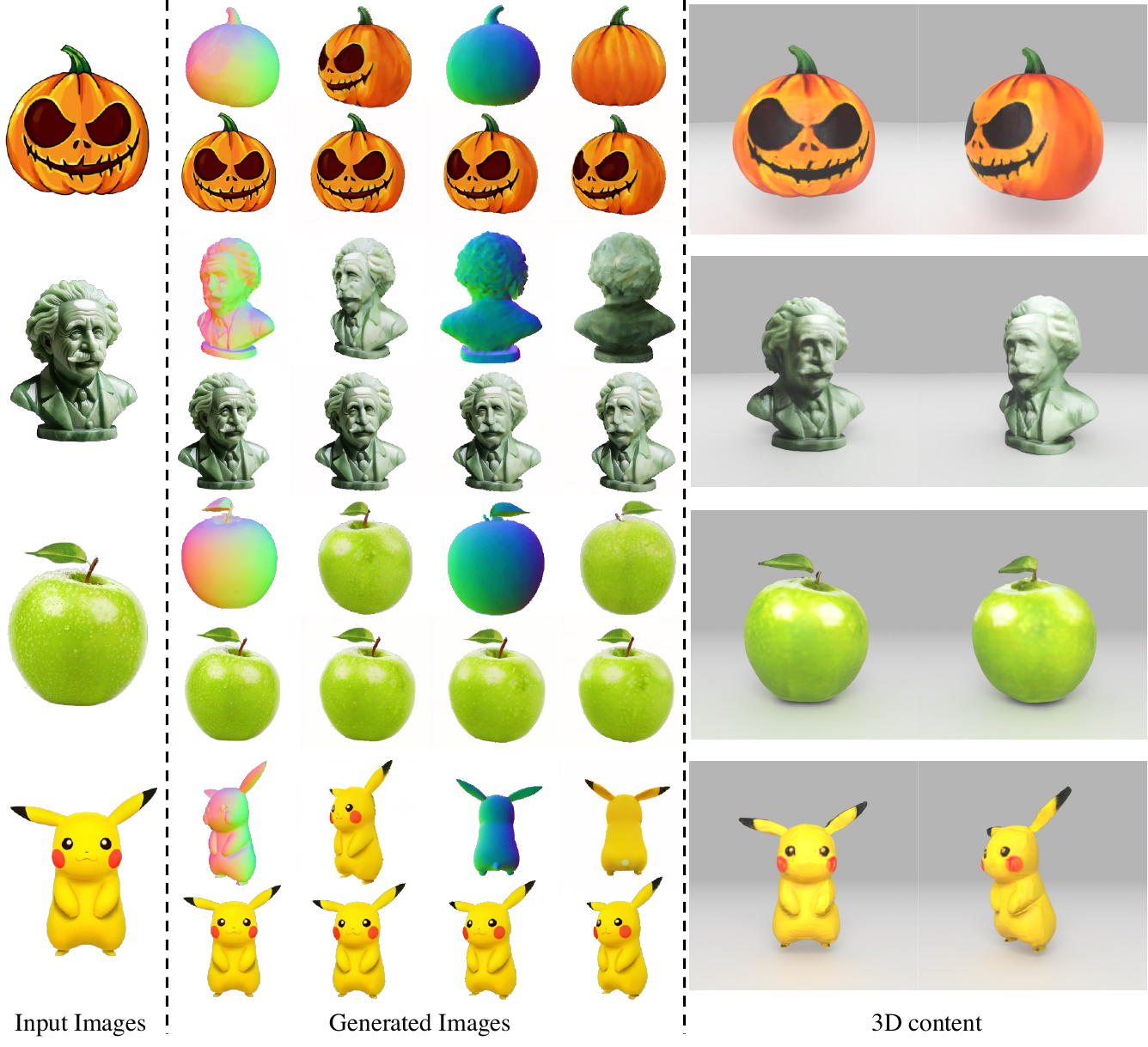}
\caption{\textbf{Envision3D generates 32 dense view images and extracts high-quality 3D content from one input image in 3-4 minutes.} 
}
\label{fig:head}	
\end{figure}
\vspace{-1.3cm}

\begin{abstract}
We present \textbf{Envision3D}, a novel method for efficiently generating high-quality 3D content from a single image. Recent methods that extract 3D content from multi-view images generated by diffusion models show great potential.
However, it is still challenging for diffusion models to generate \emph{dense} multi-view consistent images, which is crucial for the quality of 3D content extraction.
To address this issue, we propose a novel cascade diffusion framework, which decomposes the challenging dense views generation task into two tractable stages, namely anchor views generation and anchor views interpolation.
In the first stage, we train the image diffusion model to generate global consistent anchor views conditioning on image-normal pairs.
Subsequently, leveraging our video diffusion model fine-tuned on consecutive multi-view images, we conduct interpolation on the previous anchor views to generate extra dense views. 
This framework yields dense, multi-view consistent images, providing comprehensive 3D information. 
To further enhance the overall generation quality, we introduce a coarse-to-fine sampling strategy for the reconstruction algorithm to robustly extract textured meshes from the generated dense images.
Extensive experiments demonstrate that our method is capable of generating high-quality 3D content in terms of texture and geometry, surpassing previous image-to-3D baseline methods. 
GitHub repository: \url{https://github.com/PKU-YuanGroup/Envision3D}

\end{abstract}
\vspace{-1cm}

\section{Introduction}
\label{sec:intro}
Generating 3D content from a single image is essential for diverse applications such as virtual reality, gaming, and robotics.
Despite rapid advancements~\cite{mildenhall2021nerf,kerbl20233d} for 3D modeling from multi-view images, obtaining 3D content from a single image remains challenging because it requires not only reconstructing the visible parts but also inferring the invisible parts. Such a task demands that the model possesses a profound comprehension of the 3D world.

Recently, diffusion models~\cite{ho2020denoising,song2020denoising,rombach2022high} have achieved great success in 2D image generation, which opens up new opportunities for 3D generation tasks. Pioneer work DreamFusion~\cite{poole2022dreamfusion} proposes Score Distillation Sampling (SDS) to distill 2D diffusion priors into 3D representations, thus achieving text-to-3D generation tasks. Zero123~\cite{liu2023zero} proposes to fine-tune an image-to-image diffusion model with camera views as extra conditions, enabling the diffusion model 3D aware. Utilizing SDS optimization, 3D content can be generated from a single image. MVdream~\cite{shi2023mvdream} proposes a multi-view attention mechanism to generate multi-view consistent images and also implement SDS for 3D content generation. However, these SDS-based methods usually require extensive optimization iterations, costing hours to generate 3D content. Moreover, as different views are sampled in each iteration, these methods depend on the optimization process to maintain 3D consistency, leveraging the natural properties of 3D representations. The unstable optimization process could lead to low-quality 3D content, such as over-saturated textures and multi-face problems~\cite{poole2022dreamfusion}.

More recently, several works including SyncDreamer~\cite{liu2023syncdreamer} and Wonder3D~\cite{long2023wonder3d} propose to train diffusion models to directly generate multi-view consistent images at one time and implement reconstruction algorithms to extract 3D content from generated images. However, due to the limited number of multi-view consistent images, the generated 3D content usually suffers from low quality including blurred texture and distorted geometry.

In this work, our goal is to scale up the number of multi-view consistent images generated by the diffusion model, thereby providing comprehensive 3D information, which could further improve the quality of 3D content extracted by reconstruction algorithms.
However, several challenges remain unresolved. Firstly, generating more views demands that the diffusion model learn more complicated data distributions, where simply expanding existing methods could result in training non-convergence. Secondly, training efficiency needs to be improved. The prevailing multi-view diffusion models fine-tuned from image diffusion models tend to be less efficient with an increasing number of dense views. Thirdly, the dense multi-view images generated by diffusion models inevitably suffer from imperfect consistency, especially when scaling up to a large number of views. The inconsistency leads to the issue that the original 3D reconstruction algorithms may not be robust enough to extract high-quality 3D content. 

To this end, this work proposes \textbf{Envision3D}, a novel method for efficiently generating high-quality 3D content from a single image. We introduce a cascade diffusion framework, which decomposes the challenging dense views generation task into two tractable stages, namely anchor views generation and anchor views interpolation. Specifically, in the first stage, we incorporate fine-grained image-normal pairs into the diffusion model, which accelerates model convergence and facilitates the generation of anchor view images that are both semantically and geometrically consistent. In the second stage, since the video diffusion model processes multiple views efficiently and contains rich 3D prior compared to image diffusion models, we propose fine-tuning the video diffusion model conditioned on anchor views to generate additional dense views in an interpolation manner. This framework yields dense multi-view consistent images, providing comprehensive 3D information. To robustly extract high-quality textured meshes, a coarse-to-fine sampling strategy for the reconstruction algorithm is further proposed. This strategy optimizes the 3D content starting with anchor views to establish basic texture and geometry globally, then densely samples interpolation views for detail refinement, ensuring gradual and balanced enhancement of 3D quality. We evaluate our methods on GSO dataset~\cite{downs2022google} and various collected images. Envision3D demonstrates superior performance compared with competitive baseline methods. We highlight some of the results in Figure~\ref{fig:head}. 

To sum up, Envision3D contributes in the following aspects: 
\vspace{-0.3cm}
\begin{itemize}

\item A novel cascade diffusion framework decomposes the challenging dense views generation task into two tractable stages namely anchor views generation and anchor views interpolation, generating \textbf{32} consistent dense images across multiple views.

\item To improve training efficiency, several advancements are made. For the anchor views generation, we utilize image-normal pairs to speed up model convergence. For anchor views interpolation, we propose to fine-tune the video diffusion model, which efficiently processes multiple views and contains rich 3D prior compared to image diffusion models.

\item 
A novel coarse-to-fine sampling strategy robustly extracts 3D content by first optimizing the global texture and geometry starting from coarse anchor views and then refining the details through dense interpolation views.

\item Extensive experiments are conducted to evaluate the performance of our proposed method, including evaluating the GSO dataset~\cite{downs2022google} and testing with various collected images. Envision3D is capable of generating high-quality 3D content in terms of texture and geometry, surpassing previous image-to-3D baseline methods.

\end{itemize}

\section{Related Work}
\subsection{3D Generation using 2D Diffusion Models}

Recent 2D diffusion models~\cite{ho2020denoising,song2020denoising,rombach2022high,blattmann2023stable} make impressive advances in generating images from various conditions. Pioneer work DreamFusion~\cite{poole2022dreamfusion} attempts to distill prior from powerful 2D diffusion models to generate 3D content with Score Distillation Sampling (SDS). Following works ~\cite{wang2023score,wang2023prolificdreamer,seo2023ditto,yu2023points,lin2023magic3d,seo2023let,tsalicoglou2023textmesh,zhu2023hifa,huang2023dreamtime,armandpour2023re,wu2023hd,chen2023it3d,sun2023dreamcraft3d,tang2023make,melas2023realfusion,qian2023magic123,xu2022neurallift,raj2023dreambooth3d,zhang2023repaint123,shen2023anything,yu2023hifi,zhang2023avatarstudio,liu2023humangaussian,tang2023dreamgaussian,yi2023gaussiandreamer,wang2023steindreamer,liu2023unidream,qiu2023richdreamer,chen2023text,wu2024consistent3d,kwak2023vivid,cheng2023progressive3d} for text-to-3D and image-to-3D adopt this distillation pipeline to optimize various 3D representations including NeRF, mesh, SDF and Gaussian Splatting~\cite{kerbl20233d}. However, lifting 2D diffusion models for 3D generation tasks usually suffers from multi-face problems due to the lack of 3D supervision. On the other hand, SDS usually requires a long time for per-shape optimization and results are of low quality including blurred and over-saturated texture.

\subsection{Multi-view Diffusion Models}
To enable diffusion model 3D-aware, Zero123~\cite{liu2023zero} injects camera view as an extra condition to diffusion model for generating images from different views. MVdream~\cite{shi2023mvdream} proposes to replace self-attention with multi-view attention in the Unet and generate multi-view consistent images. Other works~\cite{watson2022novel,gu2023nerfdiff,deng2023nerdi,zhou2023sparsefusion,tseng2023consistent,chan2023generative,yu2023long,tewari2023diffusion,yoo2023dreamsparse,szymanowicz2023viewset,tang2023mvdiffusion,shi2023zero123++,yang2023consistnet,liu2023one,wang2023imagedream,liu2023sherpa3d,li2023sweetdreamer,ma2023geodream} share a similar idea to make diffusion models 3D-aware and improve generation consistency. However, most of these works are not designed for reconstruction methods and still require SDS loss to obtain 3D content. The recent works SyncDreamer~\cite{liu2023syncdreamer} and Wonder3D~\cite{long2023wonder3d} generate multi-view consistent 2D representation and apply reconstruction methods to obtain 3D content. However, due to the lack of comprehensive geometric information and the limited number of views, the 3D content is less satisfactory in geometry and texture. In contrast, our method can generate dense views with geometric information, resulting in high-quality 3D content in terms of texture and geometry.

\subsection{Other 3D Generation/Reconstruction Methods}

Instead of developing 2D diffusion models for 3D generation, many efforts directly model the diffusion process on various 3D representations, including point clouds~\cite{nichol2022point,zeng2022lion,luo2021diffusion,zhou20213d}, meshes~\cite{liu2023meshdiffusion,gao2022get3d,qian2024atom} and neural fields~\cite{wang2023rodin,cheng2023sdfusion,karnewar2023holofusion,kim2023neuralfield,gu2023learning,anciukevivcius2023renderdiffusion,muller2023diffrf,ntavelis2023autodecoding,jun2023shap,zhang20233dshape2vecset,gupta20233dgen,erkocc2023hyperdiffusion,chen2023single,liu2023pi3d}. However, due to the limited size of 3D dataset and the challenge to obtain ground truth 3D representations, most of the works only experiment on small scale datasets, resulting in limited performance. 

Recently, LRM~\cite{hong2023lrm} first proposes to train a deterministic model to predict tri-plane NeRF from a single image by multi-view supervision. The following works extend LRM to in-the-wild images~\cite{jang2023nvist}, multi-view images reconstruction~\cite{li2023instant3d}, combine with diffusion~\cite{xu2023dmv3d,tang2024lgm} and replacing NeRF with Gaussian Splatting~\cite{zou2023triplane,tang2024lgm}. However, it is still challenging to generate high-quality 3D content with deterministic models.

\section{Preliminaries and Problem Formulation}
\subsection{Diffusion Models}
Diffusion Models~\cite{ho2020denoising,sohl2015deep} are proposed to approximate a data distribution $q(\mathbf{x})$ by learning a probabilistic model $p_{\theta}(\mathbf{x}_{0}) = \int{p_{\theta}(\mathbf{x}_{0:T})\, \mathrm{d}\mathbf{x}_{1:T}}$. The joint distribution between $\mathbf{x}_{0}$ and random latent variables $\mathbf{x}_{1:T}$ is characterized by a reverse Markov Chain $p_\theta(\mathbf{x}_{0:T})=p(\mathbf{x}_T) \prod_{t=1}^T p_\theta(\mathbf{x}_{t-1}|\mathbf{x}_t)$, where $p_(\mathbf{x}_T)$ is a standard normal distribution and the transition kernels $p_\theta(\mathbf{x}_{t-1}|\mathbf{x}_t)=\mathcal{N}(\mathbf{x}_{t-1};\mathbf{\mu}_\theta(\mathbf{x}_t,t),\sigma^2_t \mathbf{I})$. By constructing a forward Markov Chain $q(\mathbf{x}_{1:T}|\mathbf{x}_0)=\prod_{t=1}^{T} q(\mathbf{x}_t|\mathbf{x}_{t-1})$, $\mathbf{\mu}_\theta(\mathbf{x}_t,t)$ can be defined as $\frac{1}{\sqrt{\alpha}_t}\left(\mathbf{x}_t - \frac{\beta_t}{\sqrt{1-\bar{\alpha}_t}} \mathbf{\epsilon}_\theta (\mathbf{x}_t, t)\right)$, in which the noise predictor $\mathbf{\epsilon}_\theta$ can be trained by 

\begin{equation}
    \ell=\mathbb{E}_{t,\mathbf{x}_0,\mathbf{\epsilon}}\left[\|\mathbf{\epsilon} - \mathbf{\epsilon}_\theta (\sqrt{\bar{\alpha}_t} \mathbf{x}_0+\sqrt{1-\bar{\alpha}_t}\mathbf{\epsilon}, t)\|_2\right],
\end{equation}
where $\bar{\alpha}_t$ is a constant and $\mathbf{\epsilon}$ sampled from standard normal distribution.

Based on diffusion models, the widely-used latent diffusion model~\cite{rombach2022high} maps data into a low-dimensional space with a pre-trained variational
auto-encoder (VAE)~\cite{esser2021taming} and performs the diffusion process on this latent space. This design saves computational costs and enables diffusion models to scale up to larger datasets and more complicated tasks including video generation and 3D generation. In this work, we finetune our diffusion models based on powerful latent diffusion models and their variants. 

\subsection{Problem Formulation for One Image to 3D}
\label{sec:problem}
Given one image $y$ as an input condition, our goal is to generate corresponding 3D content $X$. To make use of the powerful latent diffusion models that have strong generalization ability, instead of directly modeling 3D distribution, we formulate the 2D joint distribution from 3D content as $p(\mathbf{z})=p(x^{1:k},n^{1:k} | y)$,
where $x^{1:k}$, $n^{1:k}$ are multi-view images and normal maps respectively. These 2D representations can be generated by a diffusion model $G$ given the condition image $y$, 
\begin{equation}
    (x^{1:k},n^{1:k}) = G(y),
    \label{eq:g}
\end{equation}
Then, 3D content can be extracted from these 2D representations by applying reconstruction algorithms $R$, 

\begin{equation}
    X = R(x^{1:k},n^{1:k}),
\end{equation} 
To enable diffusion models to provide more comprehensive information on 3D content and thereby improve the quality of the extraction, we aim to scale up the number of views $k$. However, increasing $k$ makes the distribution $p(\mathbf{z})$ more complicated, which leads to non-convergence problems when training diffusion models. To address this issue, we simplify the 2D joint distribution $p(\mathbf{z})$ to two tractable distributions $p(\mathbf{z_1}), p(\mathbf{z_2})$ and propose a cascade diffusion framework $G = (G_1, G_2)$ learn them respectively. Specifically, $G_1$ generates 2D representations for $k_1$ anchor views and $G_2$ interpolates between anchor views to generate $k_2$ dense interpolation views (normal maps are omitted in this stage). Formally, our pipeline is formulated as,
\vspace{-0.5cm}

\begin{equation}
p(\mathbf{z_1}) = p(x_a^{1:k_1},n_a^{1:k_1} | y) , 
    (x_a^{1:k_1},n_a^{1:k_1}) = G_1(y),
    \label{eq:s1}
\end{equation}

\vspace{-0.5cm}

\begin{equation}
    p(\mathbf{z_2}) = p(x_i^{1:k_2} | x_a^{1:k_1}),(x_i^{1:k_2}) = G_2(x_a^{1:k_1}),
    \label{eq:s2}
\end{equation}

\vspace{-0.5cm}

\begin{equation}
    X = R(x_a^{1:k_1},n_a^{1:k_1},x_i^{1:k_2}),
    \label{eq:s3}
\end{equation} 
where the subscript $a$ and $i$ represent anchor views and interpolation views respectively. In addition to decomposing the joint distribution, this formulation allows us to utilize different diffusion models to efficiently learn the distribution at specific stages.

\begin{figure}[ht]
\vspace{-0.7cm}
\centering
\includegraphics[width=\linewidth]{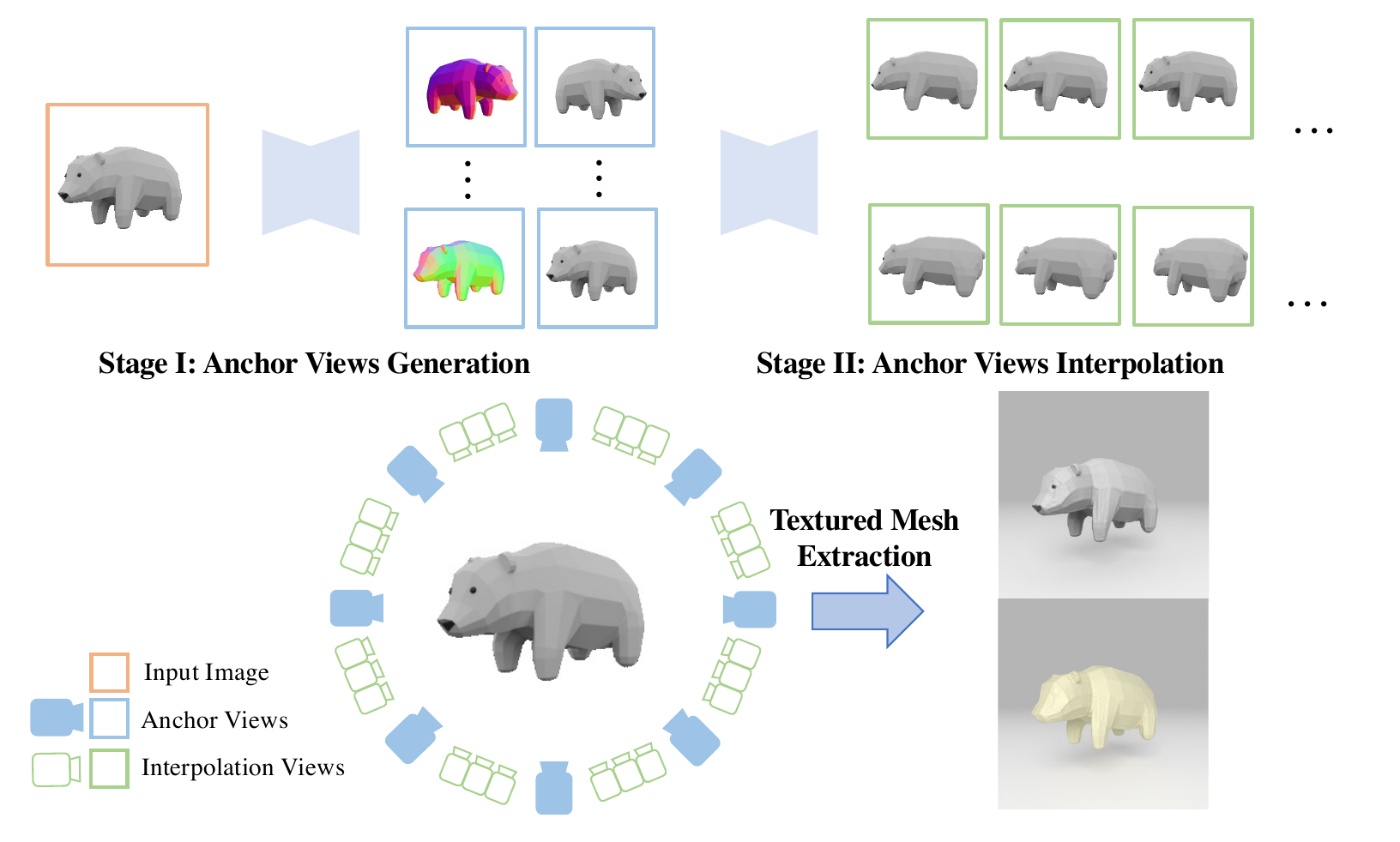}
\caption{\textbf{Overview of Envision3D.} Given an input image, Stage I first generates anchor view images with aligned normal maps. Then Stage II interpolates between previous anchor views to generate dense interpolation views. Finally, 3D content is generated through the textured mesh extraction.
}
\vspace{-1cm}
\label{fig:frame}	
\end{figure}

\section{Envision3D}
\label{sec:method}
The overview of Envision3D, based on problem formulation in Section~\ref{sec:problem}, is shown in Figure~\ref{fig:frame}. Starting from an input image, a multi-view diffusion model first generates anchor view images with aligned normal maps in Stage I, which matches Equation~\ref{eq:s1} and is described in Section~\ref{sec:s1}. Subsequently, in stage II, a pseudo-3D diffusion model fine-tuned from the video diffusion model is proposed to generate dense views by interpolating anchor views, as formulated by Euqation~\ref{eq:s2} and described in Section~\ref{sec:s2}. In Section~\ref{sec:s3}, we further introduce a coarse-to-fine reconstruction algorithm to robustly extract 3D content.

\subsection{Stage I: Anchor Views Generation}
\label{sec:s1}

\begin{figure}[ht]
\centering
\includegraphics[width=\linewidth]{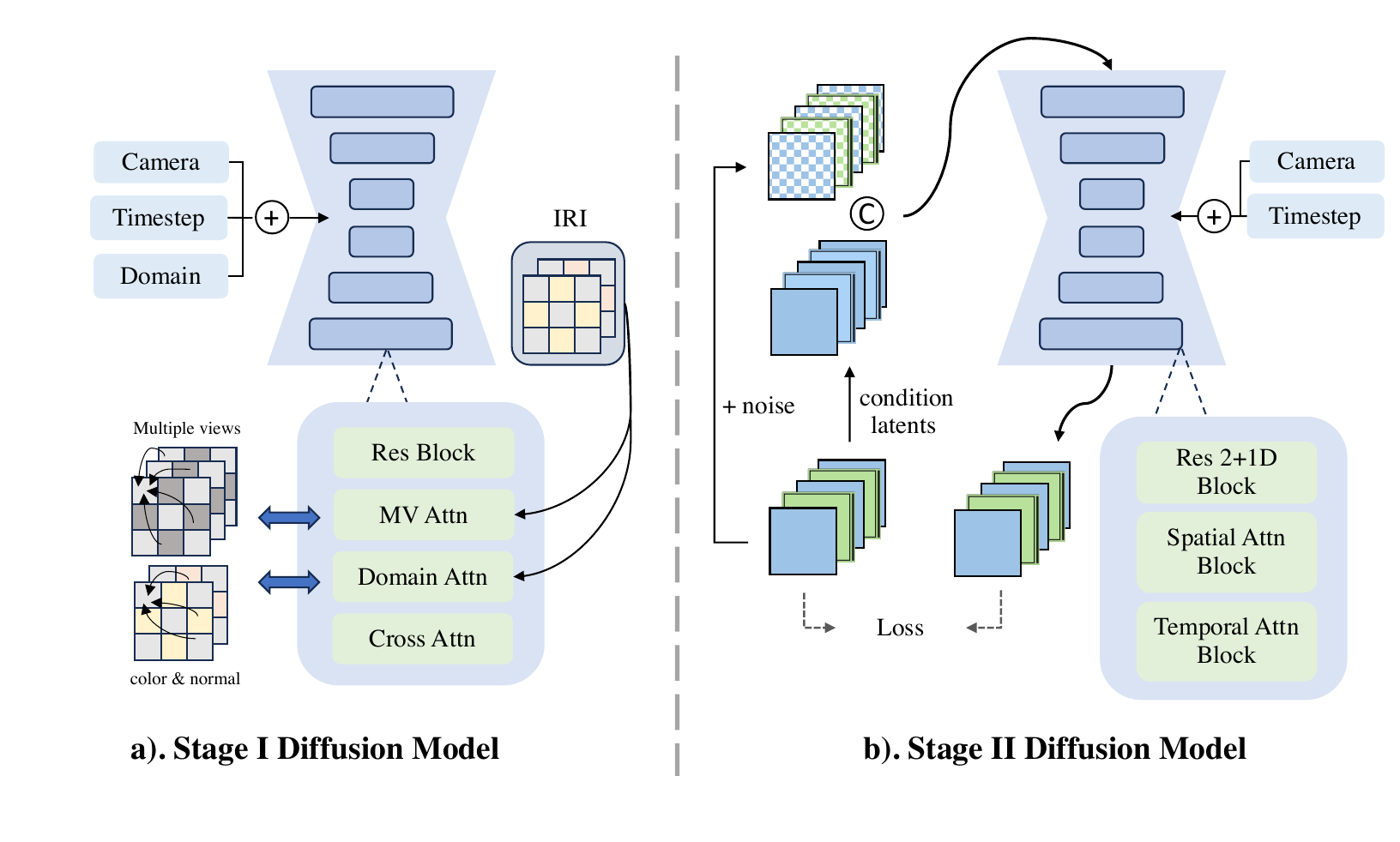}
\caption{\textbf{a) Stage I diffusion model.} We implement multi-view attention and cross-domain attention to enforce multi-view consistency and domain alignment. We propose an Instruction Representation Injection (IRI) module to inject image-normal pairs into the diffusion model. \textbf{b) Stage II diffusion model.} We fine-tune the video diffusion model composed of spatial-temporal blocks, ensuring consistency among local dense views. The conditional anchor view latents are reorganized and concatenated with noisy latents, which are taken as model input. }
\label{fig:model}	
\end{figure}

To generate consistent images across multiple anchor views, we apply a multi-view attention mechanism~\cite{shi2023mvdream}  in the diffusion model, as shown in Figure~\ref{fig:model} a). The multi-view attention modifies the original self-attention layer to include awareness of multiple views. In this manner, the diffusion model learns to understand the relationships between multiple views, enabling it to generate consistent multi-view representations. To further enable the diffusion model to generate aligned images and normal maps, we implement a cross-domain attention mechanism following Wonder3D~\cite{long2023wonder3d}.  This mechanism is designed to promote information fusion for images and corresponding normal maps. Through this cross-domain attention layer, the diffusion model establishes a strong correlation between the generation processes of both representations.

For the diffusion model to learn comprehensive 3D multi-view information, we scale up to 8 views, which are evenly distributed around the input image with the same elevation angle. However, generating 8-view consistent images with normal maps leads to slow convergence during model training. To address this issue, we propose an Instruction Representation Injection (IRI) module to inject extra conditional representations into the diffusion model, as shown in Figure~\ref{fig:model} a). Specifically, given an input image as a condition, we utilize a pre-trained normal prediction model~\cite{eftekhar2021omnidata} to generate the normal map and obtain pre-aligned image-normal conditions. After encoding by VAE, we inject these fine-grained representation pairs into multi-view attention and cross-domain attention to instruct the model generating detailed images with aligned normal maps. The main insight behind this is the diffusion model is not originally capable of predicting normal maps from images. Injecting pre-aligned image-normal pairs as instruction speeds up model convergence, improving training efficiency.

\subsection{Stage II: Anchor Views Interpolation}
\label{sec:s2}

From stage I, we obtain global semantic and geometric consistent anchor views. Recalling Equation~\ref{eq:s2}, we propose a novel method termed anchor views interpolation to generate dense interpolation views using anchor views as input conditions, which is described in Figure~\ref{fig:model} b).

In stage II, different from prior works fine-tuning from image diffusion models, our diffusion model is based on image-to-video (I2V) diffusion model~\cite{blattmann2023stable}. The design rationale behind this is grounded in a two-fold perspective. Firstly, the modality of video can be conceptualized as a series of images captured from dynamic 3D worlds. Trained on extensive video datasets, the I2V diffusion model possesses a more robust inherent understanding of 3D space, which enables the I2V diffusion model to more seamlessly adapt to dense views generation when compared to image diffusion models that typically lack 3D prior. Consequently, utilizing the I2V diffusion model as a base model offers a more efficient and effective approach for training the diffusion model that generates dense multi-view consistent images. Secondly, the I2V diffusion model employs a pseudo-3D Unet composed of spatial-temporal blocks, thus promoting content continuity and temporal consistency between video frames. Compared to previous image/text-to-3d methods that relied solely on multi-view attention to achieve multi-view consistency, this architecture provides a more effective means of maintaining consistency among local dense views.

Based on the pre-trained I2V diffusion model, we fine-tune it to generate dense views by interpolating conditional anchor views. During training, we add noise to $k$ consecutive views to obtain noisy latents, where the first, middle, and last views are designated as anchor views. We then construct the condition latents as 
\begin{equation}
    z_c = [z_{first},PADs_1,z_{middle},PADs_2,z_{last}] \in \mathbb{R}^{k \times c \times h \times w},
\end{equation}
where $PADs_1, PADs_2 \in \mathbb{R}^{(k-3)/2 \times c \times h \times w}$ are repeated by $z_{first}$ and $z_{middle}$ respectively. The noisy latents and condition latents are then concatenated along the channel dimension, which is taken as the model input. The diffusion process is parameterized and trained under EDM framework~\cite{karras2022elucidating}. At the inference stage, given $k_1$ anchor views, we reorganize them into $ \frac{k_1}{2} \times3 $ groups of views for generating dense interpolation views. Since the anchor view ensures global consistency, the generated dense interpolation views from multiple groups are also highly consistent.
 
\subsection{Textured Mesh Extraction}
\label{sec:s3}
To extract 3D content from generated anchor views and dense interpolation views, we adopt an SDF-based reconstruction method NeuS~\cite{wang2021neus}. Given $k_1$ anchor views that consist of images $x^{1:k_1}_a$ with normal maps $n^{1:k_1}_a$ and $k_2$ dense interpolation views $x^{1:k_2}_i$, we first apply a segmentation model to obtain object masks. Then we randomly sample a batch of pixels and the corresponding rays $v$ for color and normal map rendering. The optimization objective is,

\begin{equation}
    \mathcal{L} = \mathcal{L}_{image} + \mathcal{L}_{normal} + \mathcal{L}_{mask} + \mathcal{R}_{eikonal} + \mathcal{R}_{smooth} + \mathcal{R}_{sparse},
\end{equation}
where $\mathcal{L}_{image}$ is the MSE loss between rendered colors and generated images. $\mathcal{L}_{normal}$ denotes for loss between normal maps, $\mathcal{L}_{mask}$ denotes the binary cross-entropy loss between rendered masks and object masks. $\mathcal{R}_{eikonal}$, $\mathcal{R}_{smooth}$ and $\mathcal{R}_{sparse}$ are regularization terms to enhance optimization quality.

However, the generated anchor views and interpolation views inevitably have inconsistencies between individual pixels, especially when we scale up to such a large number of views. These errors may cause unstable optimization and lead to distorted, incomplete, and blurred 3D content. To tackle this challenge, we introduce a coarse-to-fine sampling strategy. Within a period of optimization steps, we initially sample rays from the anchor views during the early stages. Subsequently, after reaching certain threshold optimization steps, the ray sampling process is extended to involve all views.
\begin{equation}
    \begin{cases}
        v \in (x^{1:k_1}_a,n^{1:k_1}_a) , &step < threshold  \\
        v \in (x^{1:k_1}_a,n^{1:k_1}_a,x^{1:k_2}_i) , &step \geq threshold 
    \end{cases},
\end{equation}
 This approach is straightforward yet efficient. It begins with the optimization of 3D content using anchor views, which establishes the basic texture and geometry on a global scale. Following this, the dense interpolation views are sampled to refine and enhance the details. This method ensures a balanced optimization process, gradually improving the quality of  3D content by strategically refining the geometry and texture details.

\section{Experiments}
 
\subsection{Implementation Details}
We use a filtered Objaverse-LVIS~\cite{deitke2023objaverse} subset as the training dataset, which consists of around 30,000 objects. For each object, we first normalize it to the center of the scene and render 32 views using blenderproc~\cite{Denninger2023}. Our Stage I diffusion model is fine-tuned from the Stable Diffusion Image Variations Model following the settings in Wonder3D~\cite{long2023wonder3d}. Stage II diffusion model is fine-tuned from Stable Video Diffusion Model~\cite{blattmann2023stable} for 12,000 steps with a total batch size of 512. More training details are provided in the Supplementary Materials.
 
\subsection{Evaluation Settings}
\subsubsection{Baselines}
We adopt Zero123~\cite{liu2023zero}, Magic123~\cite{qian2023magic123}, One-2-3-45~\cite{liu2023one}, Point-E~\cite{nichol2022point}, Shap-E~\cite{jun2023shap}, SyncDreamer~\cite{liu2023syncdreamer} and Wonder3D~\cite{long2023wonder3d} as baseline methods. The implementation is either based on their open-source codes or the implementation from ThreeStudio~\cite{threestudio2023}. We briefly introduce SyncDreamer~\cite{liu2023syncdreamer} and Woner3D~\cite{long2023wonder3d} here. SyncDreamer~\cite{liu2023syncdreamer} is fine-tuned based on Zero123~\cite{liu2023zero} and generates 16 fixed views of images. The method extends the denoising network with 3D feature volumes and 3D-aware attention to enhance the multi-view consistency. Wonder3D~\cite{long2023wonder3d} is fine-tuned from the Stable Diffusion Image Variations Model with multi-view attention and cross-domain attention. This method generates 6 fixed views of images with corresponding normal maps. Both of them implement NeuS~\cite{wang2021neus} for 3D content extraction.

\begin{figure}[h]
\centering
\includegraphics[width=\linewidth]{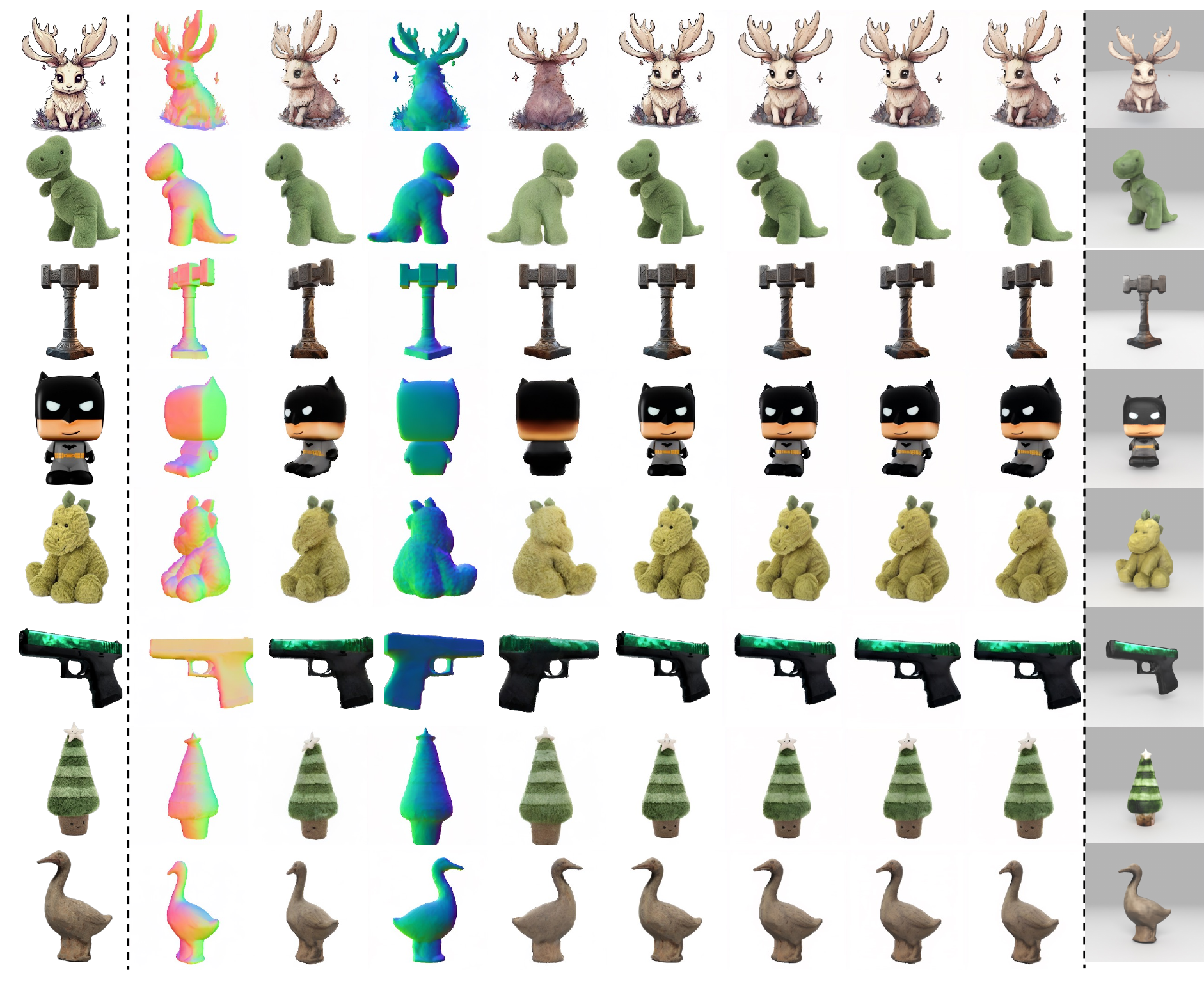}
\caption{\textbf{The qualitative results generated by Envision3D.}  We collect various images and test the performance of our method. The leftmost and rightmost column shows input images and the generated 3D content respectively.}
\label{fig:tail}	
\end{figure}

\subsubsection{Evaluation Datasets \& Metrics}
Following prior works~\cite{liu2023zero,liu2023syncdreamer,long2023wonder3d}, we evaluate our method on 30 various objects from Google Scanned Object dataset~\cite{downs2022google}. We render objects into 256$\times$256 images, which are taken as input. We also collect various images from the Internet or generated by text-to-image models to further evaluate our method. To evaluate the quality of generated and re-rendered images, we apply the metrics PSNR, SSIM~\cite{wang2004image} and LPIPS~\cite{zhang2018unreasonable}. We also adopt widely used Chamfer Distances (CD) and Volume IoU metrics for generated geometry evaluation.

\begin{figure}[h]
\centering
\includegraphics[width=\linewidth]{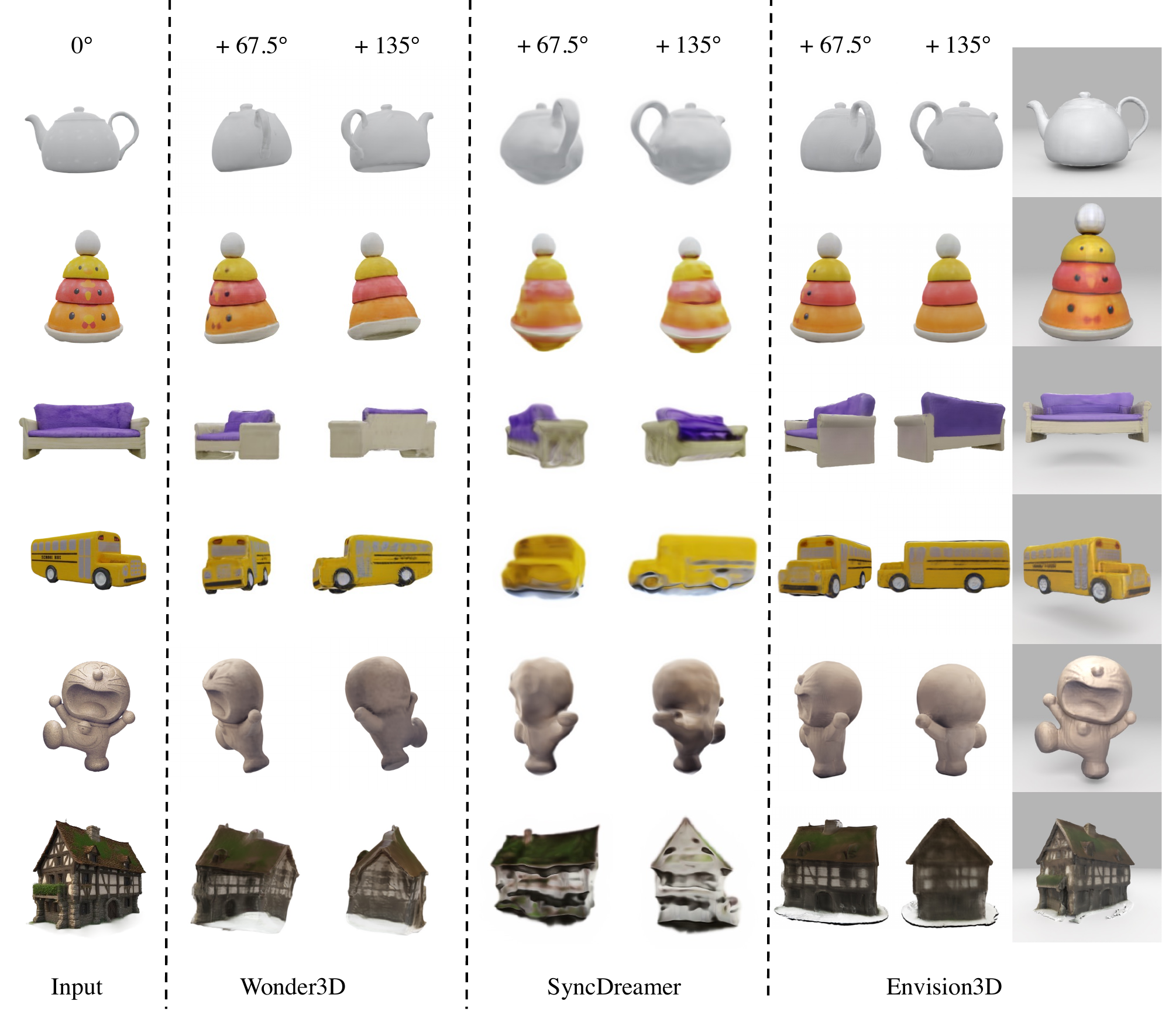}
\caption{\textbf{The qualitative comparisons with baseline methods.}  We compare re-rendered views from generated 3D content on the 4 samples from GSO dataset and 2 collected images. The rightmost column shows our generated 3D textured meshes.}
\label{fig:compare}	
\end{figure}

\subsection{Novel view synthesis}

Table~\ref{tab:nvs} presents the quantitative evaluation for synthesized views. 
Envision3D generates 32 consistent images with high quality, outperforming competitive baseline methods. Although Wonder3D achieves better metric performance, its capability to generate a large number of views is constrained. We further present qualitative results of Envision3D on various images in Figure~\ref{fig:tail}.

\subsection{3D content Generation}

As introduced in Section~\ref{sec:method}, 3D content is extracted from generated multi-view images. However, it should be noted that the quality of synthesized images does not directly reflect the quality of the 3D content. To more accurately assess the quality of the generated 3D content, we conduct an evaluation using 32 re-rendered views of the 3D content. Table~\ref{tab:nvs} demonstrates that Envision3D consistently surpasses other baseline methods by significant margins. We also provide the qualitative comparisons for re-rendered images in Figure~\ref{fig:compare}. We observe that Zero123~\cite{liu2023zero} generates over-saturated and noisy textures due to the utilization of Score Distillation Sampling. SyncDreamer~\cite{liu2023syncdreamer} struggles to produce multi-view images with high consistency, which results in blurred images upon re-rendering. Though Wonder3D~\cite{long2023wonder3d} generates consistent multi-view images, the extracted 3D content is in noisy texture due to the limited number of views. Additionally, there is a tendency for the geometry of 3D content to lean forward, which highly impacts their performance metrics. Our method clearly outperforms other competitive methods in terms of texture quality and geometry consistency. In the case of row 2, for example, the texture quality of the generated 3D content from Wonder3D is less smooth with white noise due to the sparse view. In contrast, our method generates high-quality 3D content with smooth and clean textures across views. In Figure~\ref{fig:tail}, we further test various images and show the generated 3D content in the last column. More generated 3D content and turntable animations are provided in Supplementary Materials.

\begin{table}[h]
    \centering
    \begin{tabular}{lc|ccc|ccc}
       \toprule
       Method & $\#$Views & \multicolumn{3}{c|}{Synthesized Views} & \multicolumn{3}{c}{Re-rendered Views} \\
       & & PSNR$\uparrow$ & SSIM$\uparrow$ & LPIPS$\downarrow$ & PSNR$\uparrow$ & SSIM$\uparrow$ & LPIPS$\downarrow$ \\
       \midrule
    Realfusion~\cite{melas2023realfusion}    
      &$\infty$ & 15.26 & 0.722 & 0.283  & $-$ & $-$ & $-$ \\
       Zero123(XL)~\cite{liu2023zero}    
      &$\infty$ & 18.93 & 0.779 & 0.166
      & 16.60 & 0.798 & 0.207\\
       SyncDreamer~\cite{liu2023syncdreamer}   
       &16& 20.05 & 0.798 & 0.146 & 16.02 & 0.770 & 0.249\\
       
       Wonder3D~\cite{long2023wonder3d}  
       &6& \underline{26.07} & \underline{0.924} &  \underline{0.065} & 15.89 & 0.784 & 0.214\\
       
        \textbf{Envision3d}(Ours) &32& \textbf{20.55} & \textbf{0.852} & \textbf{0.130} & \textbf{20.00} &\textbf{0.845} & \textbf{0.165} \\
       \bottomrule
    \end{tabular}
    \caption{ \textbf{The quantitative evaluation for synthesized views and re-render views.} We report PSNR, SSIM~\cite{wang2004image}, LPIPS~\cite{zhang2018unreasonable} on the GSO~\cite{downs2022google} dataset.}
    \label{tab:nvs}
\end{table}

\begin{figure}[h]

\centering
\includegraphics[width=\linewidth]{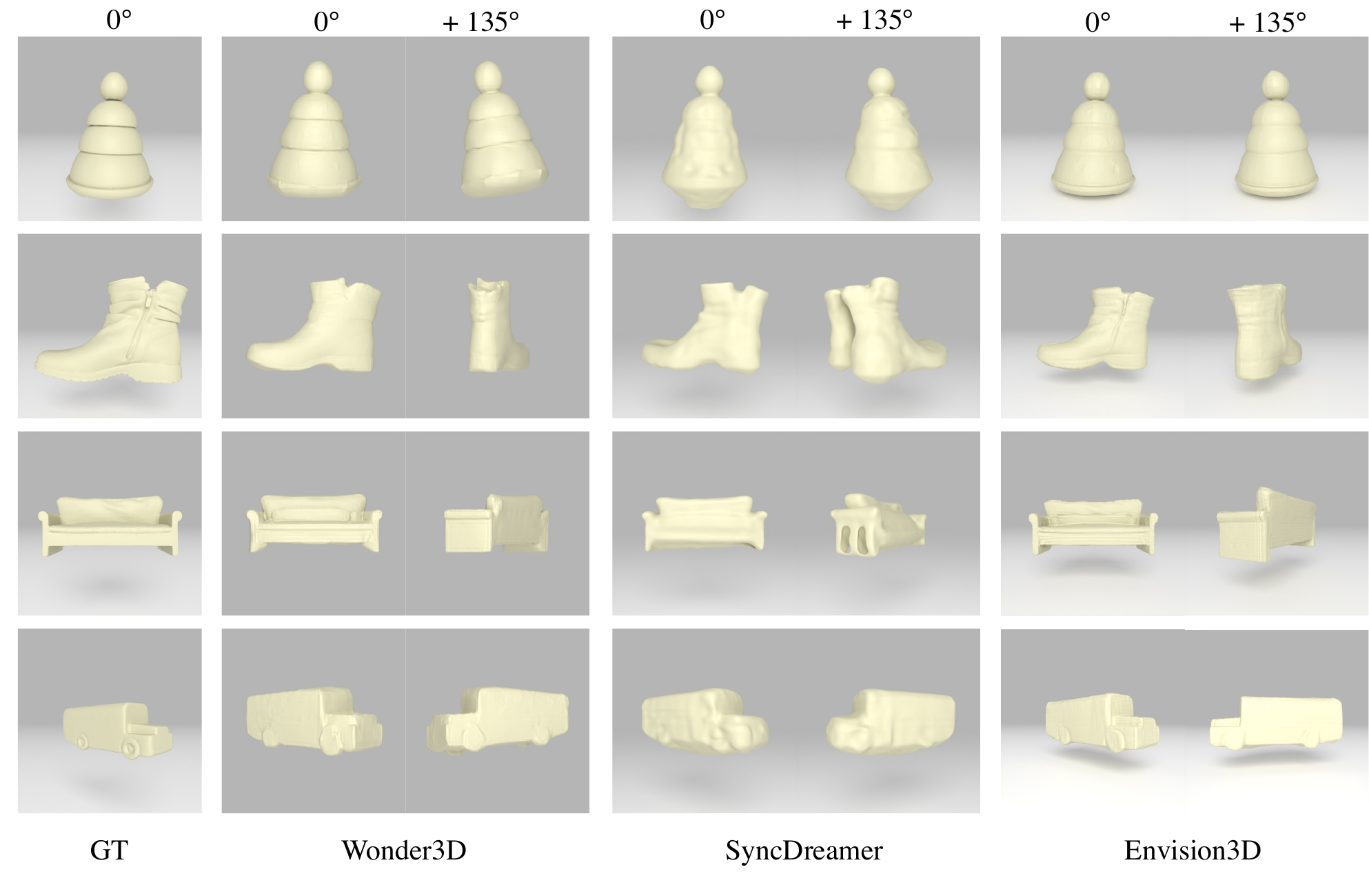}
\caption{\textbf{The qualitative comparisons with baseline methods in terms of geometry.}  We compare the generated geometry from 3D content on the GSO dataset.}
\label{fig:mesh}	
\end{figure}

\begin{table}[h]
    \centering

    \begin{tabular}{lcc|lcc}
       \toprule
       Method  & C. Dist.$\downarrow$ & V. IoU$\uparrow$ & Method  & C. Dist.$\downarrow$ & V. IoU$\uparrow$\\
       \midrule
       Magic123~\cite{qian2023magic123}
       & 0.0516 &  0.4528 & Zero123~\cite{liu2023zero}    
       & 0.0339 &  0.5035\\
       One-2-3-45~\cite{liu2023one}    
       & 0.0629 &  0.4086 &SyncDreamer~\cite{liu2023syncdreamer}   
       &  0.0261  &  0.5421\\
       Point-E~\cite{nichol2022point}    
       & 0.0426 & 0.2875 &
       Wonder3D$^*$~\cite{long2023wonder3d}  
       &  0.0253  &  0.5637 \\
       Shap-E~\cite{jun2023shap}    
       & 0.0436 &  0.3584 &
       \textbf{Envision3D}(Ours) & \textbf{0.0238} & \textbf{0.5925}\\
       \bottomrule
    \end{tabular}

    \caption{\textbf{Quantitative comparison with baseline methods.} We report Chamfer Distance and Volume IoU on the GSO~\cite{downs2022google} dataset. $^*$As Wonder3D does not provide rendering details to obtain input images, we use our rendered images to reproduce.}
    \label{tab:recon}
\end{table}

We compare the quality of the generated geometry between different methods. Table~\ref{tab:recon} shows the quantitative results, with qualitative comparisons presented in Figure~\ref{fig:mesh}. We observe that Syncdreamer~\cite{liu2023syncdreamer} tends to generate geometries that lack precision and clarity, resulting in blurred outlines and inaccuracies in shape. Wonder3D~\cite{long2023wonder3d} often generates geometries that are distorted or exhibit unintended inclinations, which detracts from the fidelity of the 3D representations. In comparison, our method distinguishes itself by incorporating image-normal pairs in the diffusion model to efficiently generate multi-view images with aligned normal maps. Followed by the robust texture mesh extraction, our method generates superior geometries from multiple views, surpassing baseline methods.  

\begin{table}[h]
    \centering
    \begin{subtable}{.5\linewidth}
      \centering
        \begin{tabular}{lccc}
           \toprule
           \# Views & PSNR$\uparrow$ & SSIM$\uparrow$ & LPIPS$\downarrow$ \\
           \midrule
           6 & 19.11 & 0.835 & 0.186 \\
           32 (Ours) & 20.00 & 0.845 & 0.165 \\
           \bottomrule
        \end{tabular}
        \caption{Effect of increasing the number of views.}
        \label{tab:ab1}
    \end{subtable}%
    \begin{subtable}{.5\linewidth}
      \centering
        \begin{tabular}{lccc}
           \toprule
            & PSNR$\uparrow$ & SSIM$\uparrow$ & LPIPS$\downarrow$ \\
           \midrule
           w/o C2F & 19.69 & 0.841 &0.170  \\
           w C2F(Ours) & 20.00 & 0.845 & 0.165 \\
           \bottomrule
        \end{tabular}
        \caption{Effect of proposed sampling strategy.}
        \label{tab:ab2}
    \end{subtable}
    \caption{\textbf{Ablation studies.} We report metrics of 32 re-rendered views.}
    \label{tab:ab_side_by_side}
\end{table}

\begin{figure}[h]
\centering
\includegraphics[width=\linewidth]{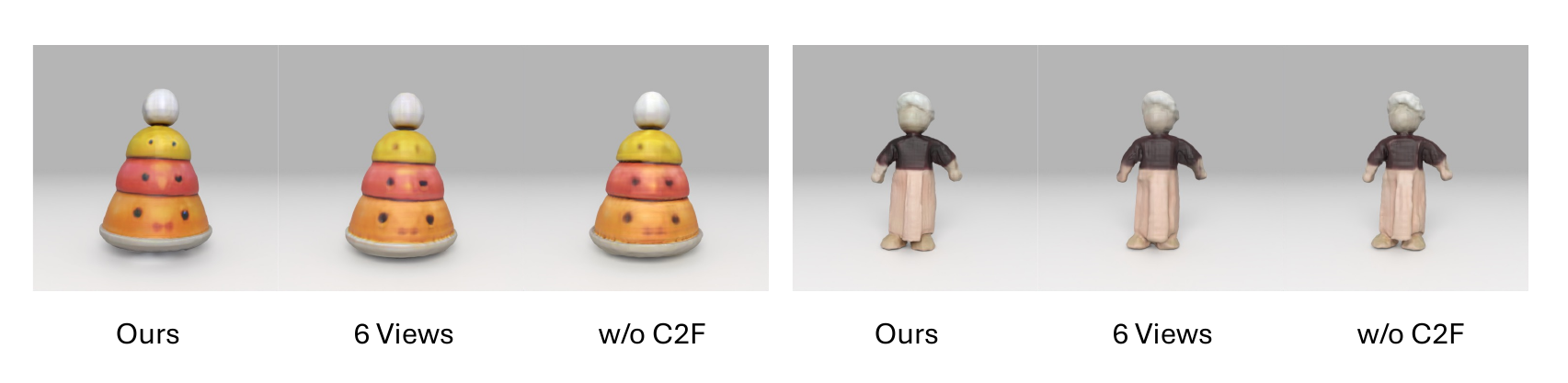}
\caption{\textbf{Qualitative ablation studies.} We show the effectiveness of using dense views for reconstruction and our proposed coarse-to-fine sampling strategy.
}
\label{fig:aba}	
\end{figure}

\subsection{Discussions}
\subsubsection{Ablation Studies}
To verify that the increased number of views enhances the quality of the generated 3D content, we conduct an ablation study. As shown in Table~\ref{tab:ab_side_by_side} (a), increasing the number of views boosts the quality of generated 3D content. It should be noted that these metric improvements are significant, as are the metric improvements in many novel methods for 3D reconstruction. To evaluate the effectiveness of the proposed coarse-to-fine (C2F) sampling strategy, we present quantitative results in Table~\ref{tab:ab_side_by_side} (b), in which our method results in better metrics. Qualitative comparisons are shown in Figure~\ref{fig:aba}. We observe that using 6 views for reconstruction, as per Wonder3D's settings, typically results in 3D content with blurred textures and imprecise geometry. Meanwhile, without our proposed C2F strategy, 3D content also lacks texture clarity and smooth geometry. In contrast, our method significantly enhances texture details and geometry smoothness of 3D content, showcasing its effectiveness.

\subsubsection{Fine-tuning Video Diffusion Model}
\begin{figure}[h]
\centering
\includegraphics[width=\linewidth]{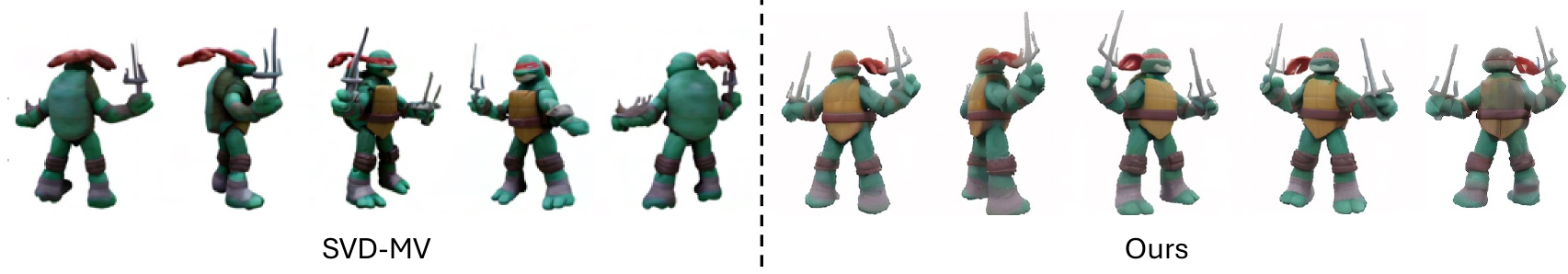}
\caption{\textbf{A qualitative result on GSO dataset.} As SVD-MV is not open-sourced, we directly copy their result in \cite{blattmann2023stable} Figure 8 for comparison. 
}
\label{fig:svd}	
\end{figure}

In the Stable Video Diffusion model~\cite{blattmann2023stable}, the authors directly fine-tune the video diffusion model on object datasets to generate 25 dense views, named SVD-MV. However, this approach suffers two major limitations. Firstly, relying solely on the network architecture consisting of spatial and temporal blocks is not suitable for generating long-term multi-view consistent images. As a result, the produced multi-view images resemble a dynamic object in multiple views rather than maintaining consistency over different views. Additionally, the efficiency of the denoising network is compromised when processing a large number of dense views simultaneously. In our method, we first generate global consistent anchor views and then utilize the fine-tuned video diffusion model for generating local dense views in groups, maintaining consistency across all views. To demonstrate the effectiveness of this design, we present qualitative results as shown in Figure~\ref{fig:svd}. The object in SVD-MV, especially the held weapons, disappears in different views and exhibits a lack of consistency. In contrast, our method successfully maintains long-term multi-view consistency, highlighting its superiority. 

\section{Conclusion}

In this work, we present Envision3D, a novel method for efficiently generating high-quality 3D content from a single image. We propose a novel cascade diffusion framework, which decomposes the challenging dense views generation task into two tractable stages with specialized diffusion models for generating and interpolating anchor views, coupled with a coarse-to-fine sampling strategy for 3D content extraction. Extensive experiments show that our method is capable of generating 3D content with high-quality texture and geometry, surpassing competitive baseline image-to-3D methods.

\clearpage  

%
%
\bibliographystyle{splncs04}
\bibliography{main}
\end{document}